\documentclass{svmult}
\usepackage{graphicx}
\usepackage{psfrag}
\usepackage{floatflt}

\begin{document}
\title*{Non-singular assembly mode changing trajectories in the workspace for the 3-R\underline{P}S parallel robot}
\author{D. Chablat \and R. Jha \and F. Rouillier \and G. Moroz}
\institute{%
  D. Chablat and R. Jha \at
  Institut de Recherche en Communications et Cybernétique de Nantes, \\
  \email{\{Damien.Chablat and Ranjan.Jha\}@irccyn.ec-nantes.fr}
  \\
  F. Rouillier \at
  INRIA Paris-Rocquencourt Institut de Mathématiques de Jussieu (UMR 7586),  \\
	\email{Fabrice.Rouillier@inria.fr} \\
	 G. Moroz \at
	INRIA Nancy-Grand Est,   \email{Guillaume.Moroz@inria.fr}}
\titlerunning{Non-singular assembly mode changing trajectories in the workspace \ldots}
\maketitle
\abstract{Having non-singular assembly modes changing trajectories for the 3-R\underline{P}S parallel robot  is a well-known feature. The only known solution for defining such trajectory is  to encircle a cusp point in the joint space. 
In this paper, the aspects and the characteristic surfaces are computed for each operation mode to define the uniqueness of the domains. Thus, we can easily see in the workspace that at least three assembly modes can be reached for each operation mode. To validate this property, the mathematical analysis of the determinant of the Jacobian is done. The image of these trajectories in the joint space is depicted with the curves associated with the cusp points.}
\section{Introduction}
\label{sec:introduction}
When designing a robot, the last step is the trajectory planning. The task of the robot is generally defined in the workspace  whereas the control loop depends on the joint space parameters. While defining the home pose of the robot, the Cartesian pose and the Joint values of the actuators are known. 
If the trajectory planning is done in the workspace by analyzing only the determinant of the Jacobian, we can reach a Cartesian pose different from the home pose but with the same joint value. This feature is called a non-singular assembly mode changing trajectory and stands only for the parallel robot.

For such robots, the inverse and direct kinematic problem (DKP) can have several solutions. To cope up with this problem, the notion of aspects was introduced for the serial robot in \cite{Borrel:1986} and for the parallel robot in \cite{Wenger:1997} and \cite{Chablat:1998}. For the serial robots, the aspects are defined as the maximal singularity-free sets in the joint space whereas in case of parallel robots, the aspects are defined as the maximal singularity-free sets in the workspace or the cross-product of the joint space by the workspace. However, there exists robots, referred  as cuspidal robots, which are able to change the inverse kinematic solution without passing through a singularity for serial robots or direct kinematic solution without passing through a singularity for parallel robots \cite{Innocenti:1992, Chablat:2001, Caro:2012, Husty:2013, Macho:2008}. The uniqueness domains are the connected subsets of the aspects induced by the {\em characteristic surface}. These notions are defined more precisely in sections 2.3 and 2.4 .

The paper elucidates the non-singular assembly mode changing trajectories in the workspace for the 3-R\underline{P}S parallel robot. In Section 2.1 we describe the 3-R\underline{P}S parallel robot, in section 2.2 we set the related kinematic equations while in section 2.3 we define the aspects for an operation mode. In section 2.4 we analyze the characteristic surfaces for an operation mode, and in section 2.5 we report the non-singular assembly modes changing trajectory between the two basic regions.
\section{Kinematics}
\subsection{Mechanism under study}
\begin{floatingfigure}[r]{37mm}
  \begin{center}
				\psfrag{A1}{$A_1$}
				\psfrag{A2}{$A_2$}
				\psfrag{A3}{$A_3$}
				\psfrag{B1}{$B_1$}
				\psfrag{B2}{$B_2$}
				\psfrag{B3}{$B_3$}
				\psfrag{P}{$P$}
				\psfrag{O}{$O$}
				\psfrag{g}{$g$}
				\psfrag{h}{$h$}
				\psfrag{x}{$x$}
				\psfrag{y}{$y$}
				\psfrag{z}{$z$}
				\psfrag{u}{$u$}
				\psfrag{v}{$v$}
				\psfrag{w}{$w$}
				\psfrag{R1}{$\rho_1$}
				\psfrag{R2}{$\rho_2$}
				\psfrag{R3}{$\rho_3$}	
        \includegraphics[scale=0.29]{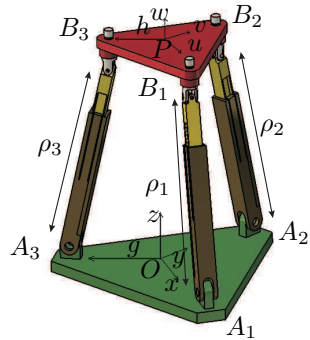}
        \caption{3-R\underline{P}S parallel robot}
        \protect\label{figure:robot}
  \end{center}
\end{floatingfigure}
The robot under study is the 3-R\underline{P}S parallel robot with three degrees of freedom. It been studied by many researchers \cite{Husty:2013,Babu:2013}.  It is the assembly of two equilateral triangles (the base and the moving platform) by three identical RPS legs where R is a revolute passive joint, \underline{P} an  prismatic joint and S a passive spherical joint. Thus, the revolute joint is connected to the fixed base and the spherical joint to the mobile platform. 

Considering the 3-R\underline{P}S parallel manipulator, as shown in figure~\ref{figure:robot}, the fixed base consists of an equilateral triangle with vertices $A_1$, $A_2$ and $A_3$, and circumradius $g$. The moving platform is another equilateral triangle with vertices $B_1$, $B_2$ and $B_3$, circumradius $h$ and circumcenter $P$. 
The two design parameters $g$ and $h$ are positive numbers. Connecting each of the pairs of vertices of $A_i$, $B_i$ ($i=1, 2, 3)$ by a limb, a rotational joint lies at $A_i$ and a spherical joint lies at $B_i$. $\rho_i$ denotes the length of each limb and their adjustment is done through an actuated prismatic joint. Thus we get five parameters, namely $g$, $h$, $\rho_1$, $\rho_2$ and $\rho_3$. $g$ and $h$ are the two design parameters determine the design of the manipulator whereas the joint parameters $\rho_1$, $\rho_2$ and $\rho_3$ determine the motion of the robot. To simplify the equations, we will study a unit robot with $g=h=1$.
\subsection{Kinematic equations}
The transformation from the moving frame to the fixed frame can be described by a position vector ${\bf p} = OP$ and a $3 \times3$ rotation matrix $\bf R$. Let $\bf u$, $\bf v$ and $\bf w$ be the three unit vectors defined along the axes of the moving frame, then the rotation matrix can be expressed in terms of the coordinates of $\bf u$, $\bf v$ and $\bf w$ as:
\begin{equation}
{\bf R}=
 \left[ \begin {array}{ccc} 
u_x & v_x & w_x \\
u_y & v_y & w_y \\
u_z & v_z & w_z 
\end {array} \right] 
\end{equation}
The vertices of the base triangle and mobile platform triangle are
\begin{equation}
{\bf A}_1= \left[ \begin {array}{c} 
g\\ 
0\\ 
0
\end {array} \right] \quad
{\bf A}_2= \left[ \begin {array}{c} 
-g/2\\
 g\sqrt {3}/2\\ 
0
\end {array} \right] \quad
{\bf A}_3= \left[ \begin {array}{c} 
-g/2\\
-g\sqrt {3}/2\\ 
0
\end {array} \right]
\end{equation}

\begin{equation}
{{\bf b}_1}= \left[ \begin {array}{c} 
h\\ 
0\\ 
0
\end {array} \right] \quad 
{{\bf b}_2}= \left[ \begin {array}{c} 
-h/2\\
 h\sqrt {3}/2\\ 
0
\end {array} \right] \quad
{{\bf b}_3}= \left[ \begin {array}{c} 
-h/2\\
-h\sqrt {3}/2\\ 
0
\end {array} \right]
\end{equation}
The coordinates of ${\bf b}_i$ with respect to fixed frame reference are obtained by ${\bf B}_i= {\bf P+R b}_i$ for $i=1,2,3$. Also the coordinates of the centre of the mobile platform in the fixed reference is ${\bf P}= [x~y~z]^T$. The distance constraints yields: 
\begin{equation}
    \label{eq:distance}
||{\bf A}_i - {\bf B}_i||= \rho_i^2  \quad {\rm with} \quad i=1, 2, 3
\end{equation}
As $A_i$ are revolute joints, the motion of the $B_i$ are constrained in planes.
This leads to the three constraint equations:
\begin{eqnarray}
    u_y h + y &=&0 \label{eq:y1}\\
    y-u_y h/2+\sqrt{3}v_y h/2 +\sqrt{3}x-\sqrt{3} u_x h/2 +3 v_x h/2 &=& 0 \label{eq:y2}\\
    y-u_y h/2-\sqrt{3}v_y h/2 -\sqrt{3}x+\sqrt{3} u_x h/2 +3 v_x h/2 &=&0 \label{eq:y3}
\label{eq:constraint}
\end{eqnarray}
Solving with respect to $x$ and $y$ we get:
\begin{eqnarray}
    y &=& -h u_y \label{eq:x}\\
    x &=&  h \left(\sqrt{3} u_x-\sqrt {3} v_y-3 u_y+3 v_x \right) \sqrt {3}/6 \label{eq:y}
\end{eqnarray}          
The coefficients of the rotation matrix can be represented by quaternions. The quaternion representation is used for modeling the orientation as quaternions do not suffer from singularities as Euler angles do. 
The quaternion rotation matrix for the parallel robot is then
\begin{equation}
    \label{eq:quaternions}
{\bf R}= \left[ \begin {array}{ccc} 
 2 {q_1}^{2}  +2 {q_2}^{2}-1~&~
-2 {q_1} {q_4}+2 {q_2} {q_3}~&~
 2 {q_1} {q_3}+2 {q_2} {q_4}\\ 
 2 {q_1} {q_4}+2 {q_2} {q_3}~&~
 2 {q_1}^{2}  +2 {q_3}^{2}-1~&~
-2 {q_1} {q_2}+2 {q_3} {q_4}\\ 
-2 {q_1} {q_3}+2 {q_2} {q_4}~&~
 2 {q_1} {q_2}+2 {q_3} {q_4}~&~
 2 {q_1}^{2}  +2 {q_4}^{2}-1
  \end {array} \right] 
\end{equation}
with $q_1^2+q_2^2+q_3^2+q_4^2 = 1$. In Equations~\ref{eq:distance},~\ref{eq:y2},~\ref{eq:y3}, we substitute $x, y$ using relations~\ref{eq:x} and \ref{eq:y}, and $\bf u, \bf v, \bf w$ by quaternion expressions using \ref{eq:quaternions}. Then (\ref{eq:y2}) and (\ref{eq:y3}) become $q_1 q_4=0$.
Thus, we have  either $q_1=0$ or  $q_4=0$. This property is associated with the notion of operation mode \cite{Caro:2013}.

\begin{figure}[!hb]
  \begin{center}
    \begin{center}
    \begin{tabular}{cc}
       \begin{minipage}[t]{57 mm}
				\psfrag{q2}{$q_2$}
				\psfrag{q3}{$q_3$}
				\psfrag{q4}{$q_4$}
        \includegraphics[scale=0.22]{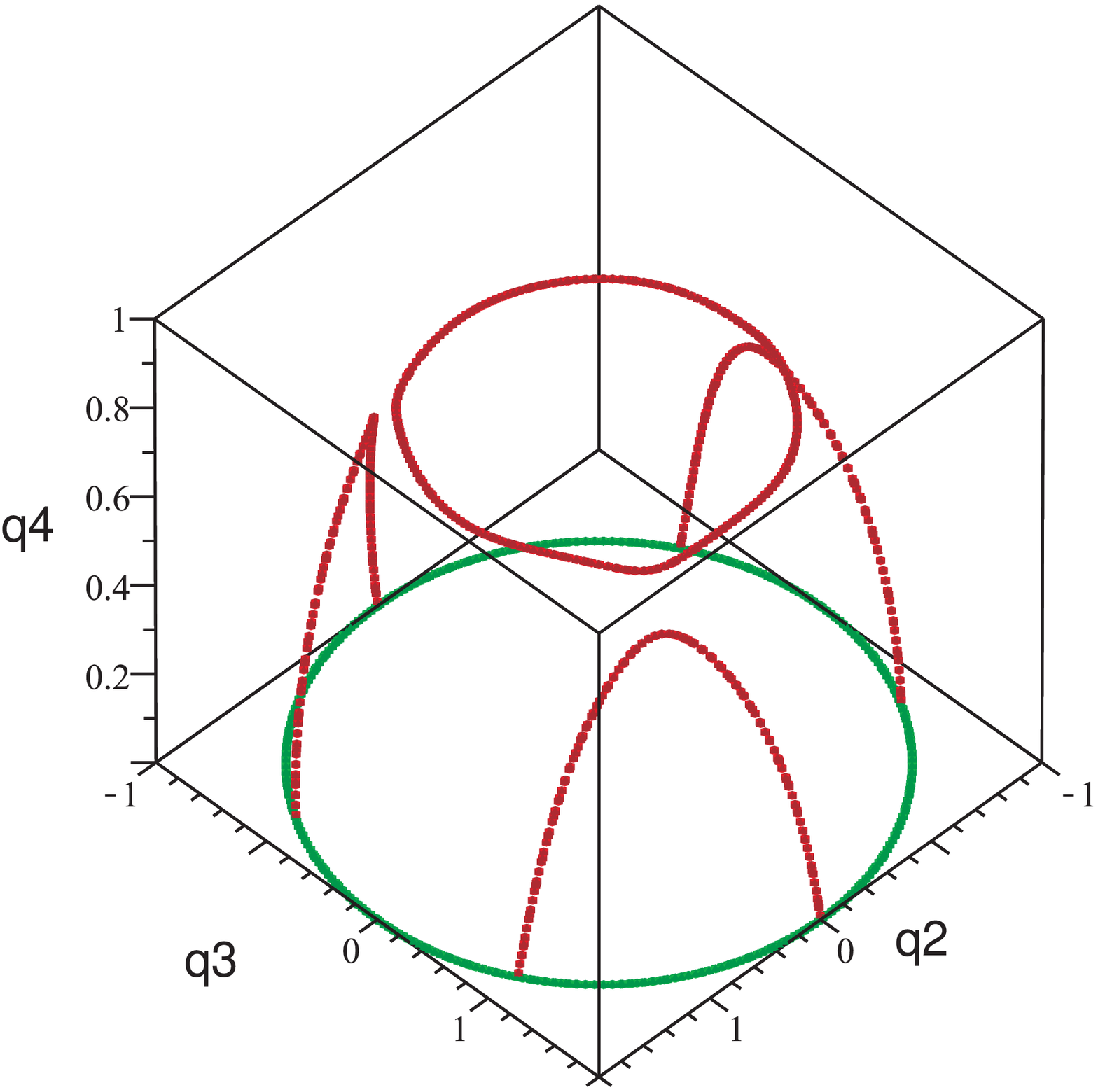}(a)
       \end{minipage} &
       \begin{minipage}[t]{57 mm}
				\psfrag{q1}{$q_1$}
				\psfrag{q2}{$q_2$}
				\psfrag{q3}{$q_3$}
        \includegraphics[scale=0.22]{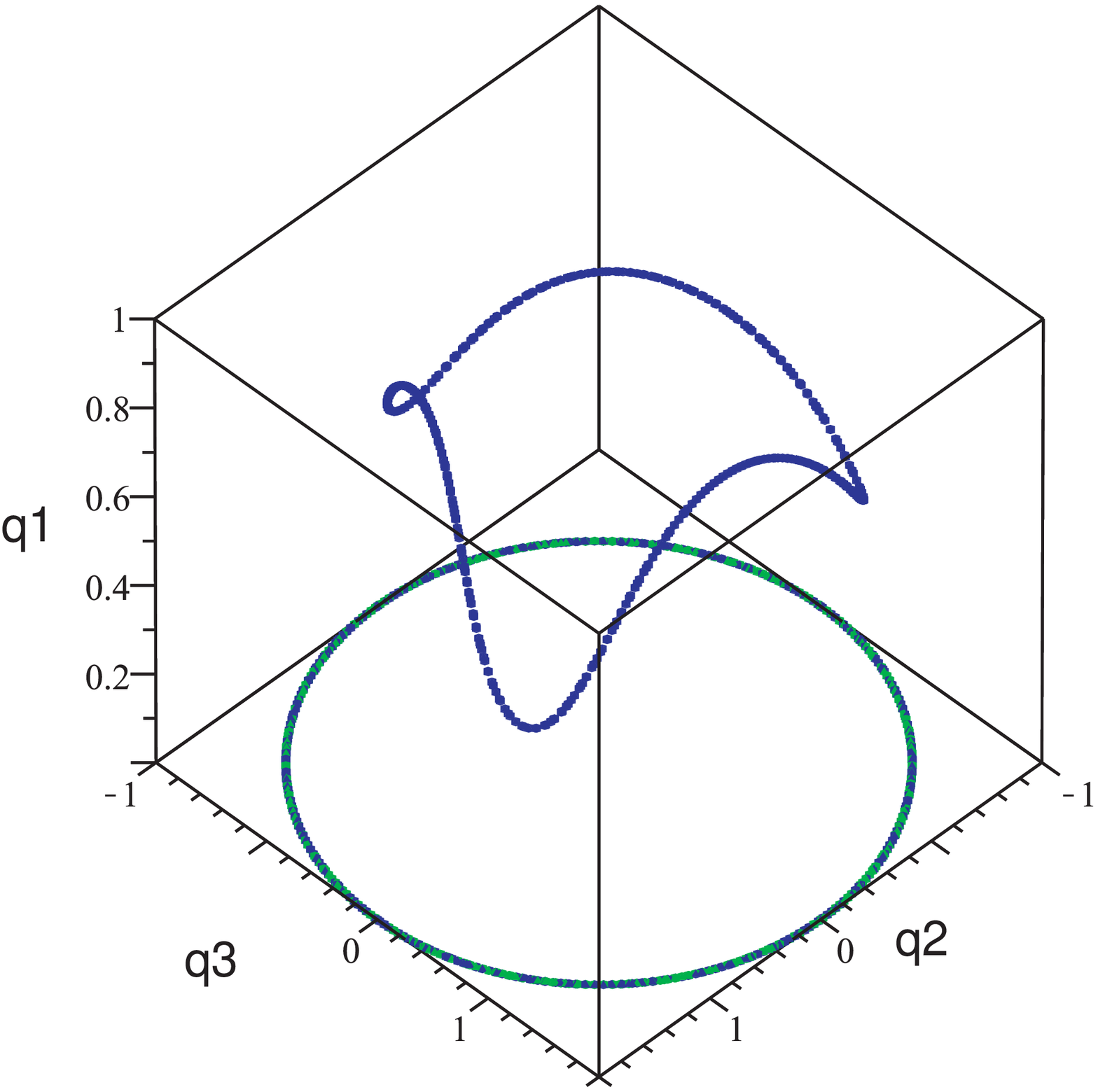}(b)
        \end{minipage}
    \end{tabular}
    \end{center}
    \caption{Singularity curves for $z=3$, $q_1=0$ (a) and $q_4=0$ (b)}
    \protect\label{figure:singularity_curves_q1_q4}
  \end{center}
\end{figure}

The notion of operation  mode (OM) was introduced in \cite{Zlatanov:2002} to explain the behavior of the DYMO robot. An operation mode is associated with a specific type of motion. For the DYMO, we have 5 operation modes: translational, rotational, planar (2 types) and mixed motions. 
In the workspace $W$, for each motion type, the $W_{OM_j}$ is defined such that
\begin{itemize}
  \item $W_{OM_j} \subset W$ 
	\item $\forall X \in W_{OM_j}$, OM is constant
\end{itemize}
For a parallel robot with several operating modes, the pose can be defined by fixing the control parameters. 
For an operation mode  $OM_j$, if we have a single inverse kinematic solution, we can then define an application that maps $\bf X$ onto $\bf q$:
\begin{equation}
    g_j({\bf X}) = {\bf q} 
\end{equation}
Then, the images in $W_{OM_j}$ of a posture $\bf q$ in the joint space $Q$ is defined by:
\begin{equation}
    g_j^{-1}({\bf q})={\bf X} \mid ({\bf X, q})\in OM_j
\end{equation}
where $g_j^{-1}$ is the direct kinematic problem restricted to the operation mode $j$.
Differentiating with respect to time the constraint equations leads to the velocity model:
\begin{equation}
  {\bf A} \dot{\bf t}+ {\bf B} \dot{\bf q}=0
\end{equation}
where $\bf A$ and $\bf B$ are the parallel and serial Jacobian matrices respectively, $\dot{\bf t}$ is the velocity of $P$ and $\dot{\bf q}$ is the joints velocity. The parallel singularities occur whenever ${\rm det}({\bf A})=0$. Let $OM_1$ (reps. $OM_2$) be the operation mode where $q_1=0$ (reps. $q_4=0$). Then ${\cal S}_{OM_1}$ and ${\cal S}_{OM_2}$ are the loci of the parallel singularities and are characterized by:
\begin{eqnarray}
{\cal S}_{OM_1}: q_4  ( 8 q_2 q_3^2q_4^6+2 q_2 q_4^{8}-64 z q_3^6q_4-96 z q_3^4q_4^3-36 z q_3^2q_4^5-6 z q_4^7\nonumber \\
-24 z^2q_2 q_3^2q_4^2-6 z^2q_2 q_4^4-32 q_2 q_3^2q_4^4-10 q_2 q_4^6+2 z^3q_4^3+96 z q_3^4q_4\nonumber \\
+72 z q_3^2q_4^3+23 z q_4^5+16 z^2q_2 q_3^2+10 z^2q_2 q_4^2+8 q_2 q_4^4-z^3q_4-36 z q_3^2q_4\nonumber \\
-21 z q_4^3-4 z^2q_2+4 z q_4 ) =0
\end{eqnarray}
\begin{eqnarray}
{\cal S}_{OM_2}: q_1^2 ( 6 q_1^7q_3+8 q_1^5q_3^3-2 z q_1^6+36 z q_1^4q_3^2+96 z q_1^2q_3^4+64 z q_3^6\nonumber \\
-18 z^2q_1^3q_3-24 z^2q_1 q_3^3 -18 q_1^5q_3-16 q_1^3q_3^3+2 z^3q_1^2+3 z {{ q_1}}^4-72 z q_1^2q_3^2\nonumber \\
-96 z q_3^4+18 z^2q_1 q_3+12 q_1^3q_3-z^3+3 z q_1^2+36 z q_3^2-4 z) =0
\end{eqnarray}
The serial singularities occur whenever $\rho_1 \rho_2 \rho_3 = 0$. The common coordinates for both operation modes are $z$, $q_2$ and $q_3$. Due to the redundancy of the quaternion representation, there exists two triplets defined by these three coordinates that represent the same pose in the same operation mode. To overcome this problem, we set $q_1>0$ and $q_4>0$. We can then depict a slice of this hypersurface by fixing one parameter as shown in Figure~\ref{figure:singularity_curves_q1_q4}. 
\subsection{Aspect for an operation mode}
\begin{figure}[ht]
    \begin{center}
    \begin{tabular}{cc}
       \begin{minipage}[t]{57 mm}
				\psfrag{z}{$z$}
				\psfrag{q2}{$q_2$}
				\psfrag{q3}{$q_3$}
				\includegraphics[scale=0.24]{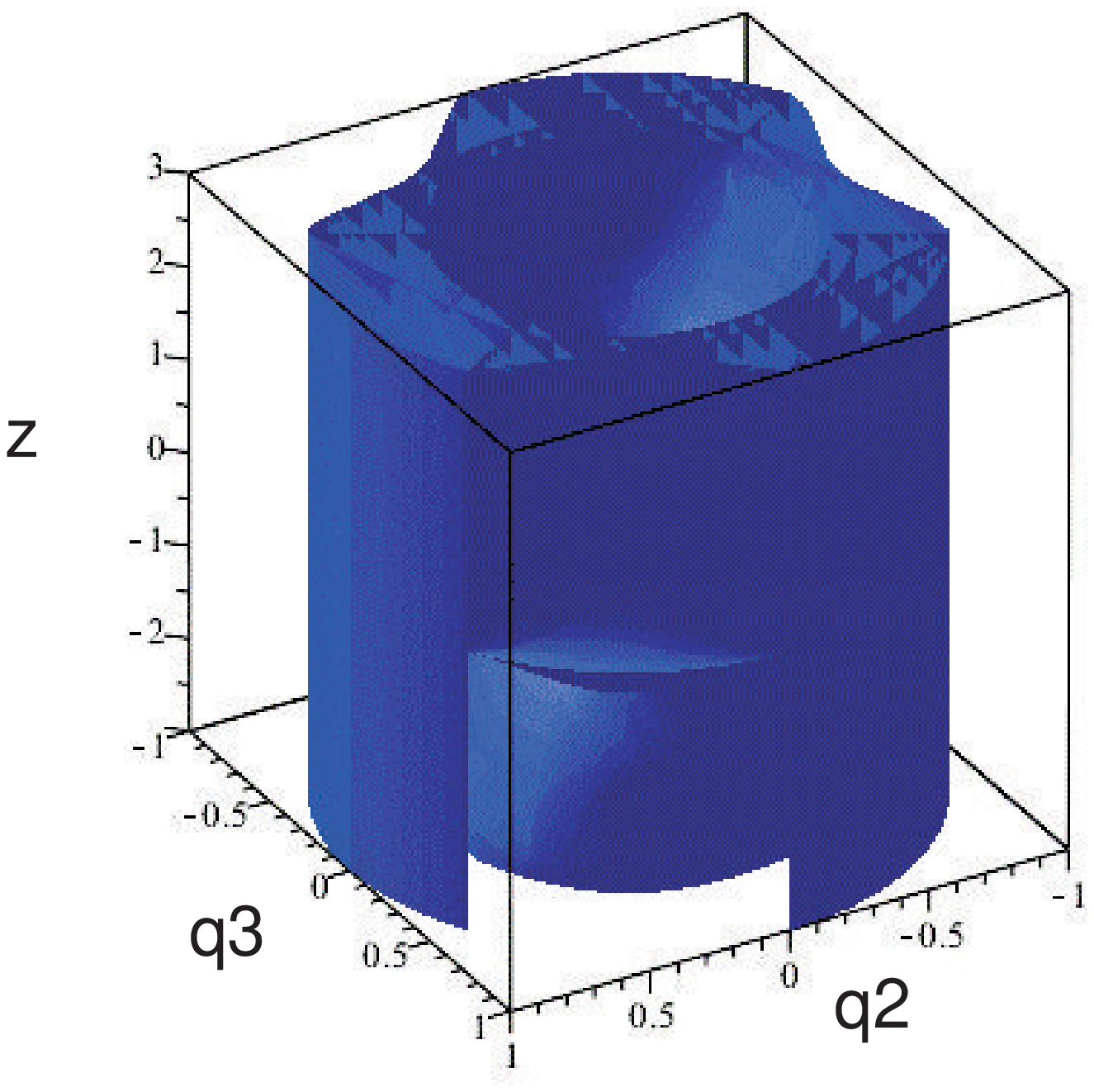}(a)
       \end{minipage} &
       \begin{minipage}[t]{57 mm}
				\psfrag{z}{$z$}
				\psfrag{q2}{$q_2$}
				\psfrag{q3}{$q_3$}
				\includegraphics[scale=0.24]{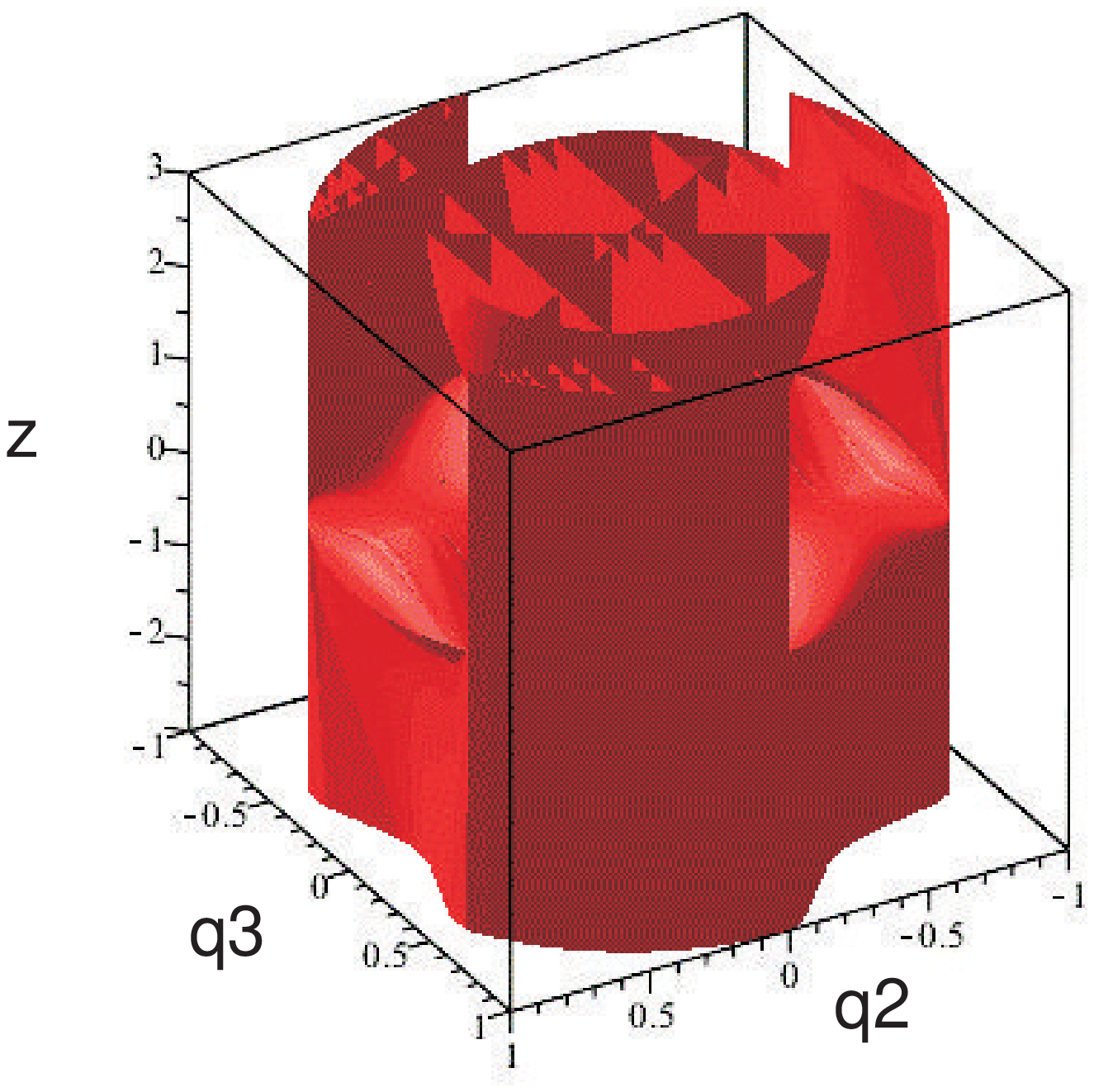}(b)
       \end{minipage}
    \end{tabular}
    \begin{tabular}{cc}
       \begin{minipage}[t]{57 mm}
				\psfrag{z}{$z$}
				\psfrag{q2}{$q_2$}
				\psfrag{q3}{$q_3$}
				\includegraphics[scale=0.24]{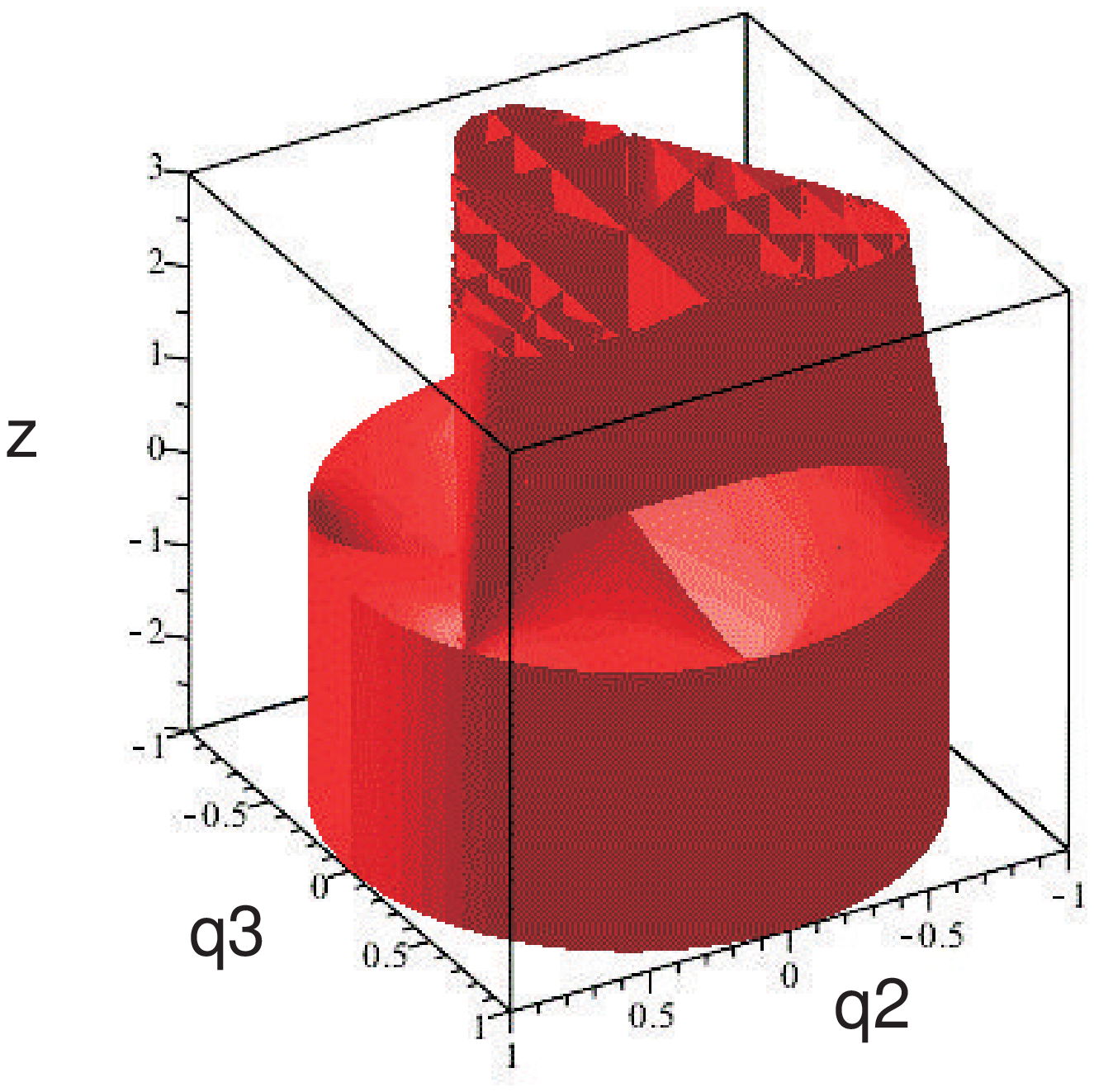}(c)
       \end{minipage} &
       \begin{minipage}[t]{57 mm}
				\psfrag{z}{$z$}
				\psfrag{q2}{$q_2$}
				\psfrag{q3}{$q_3$}
				\includegraphics[scale=0.24]{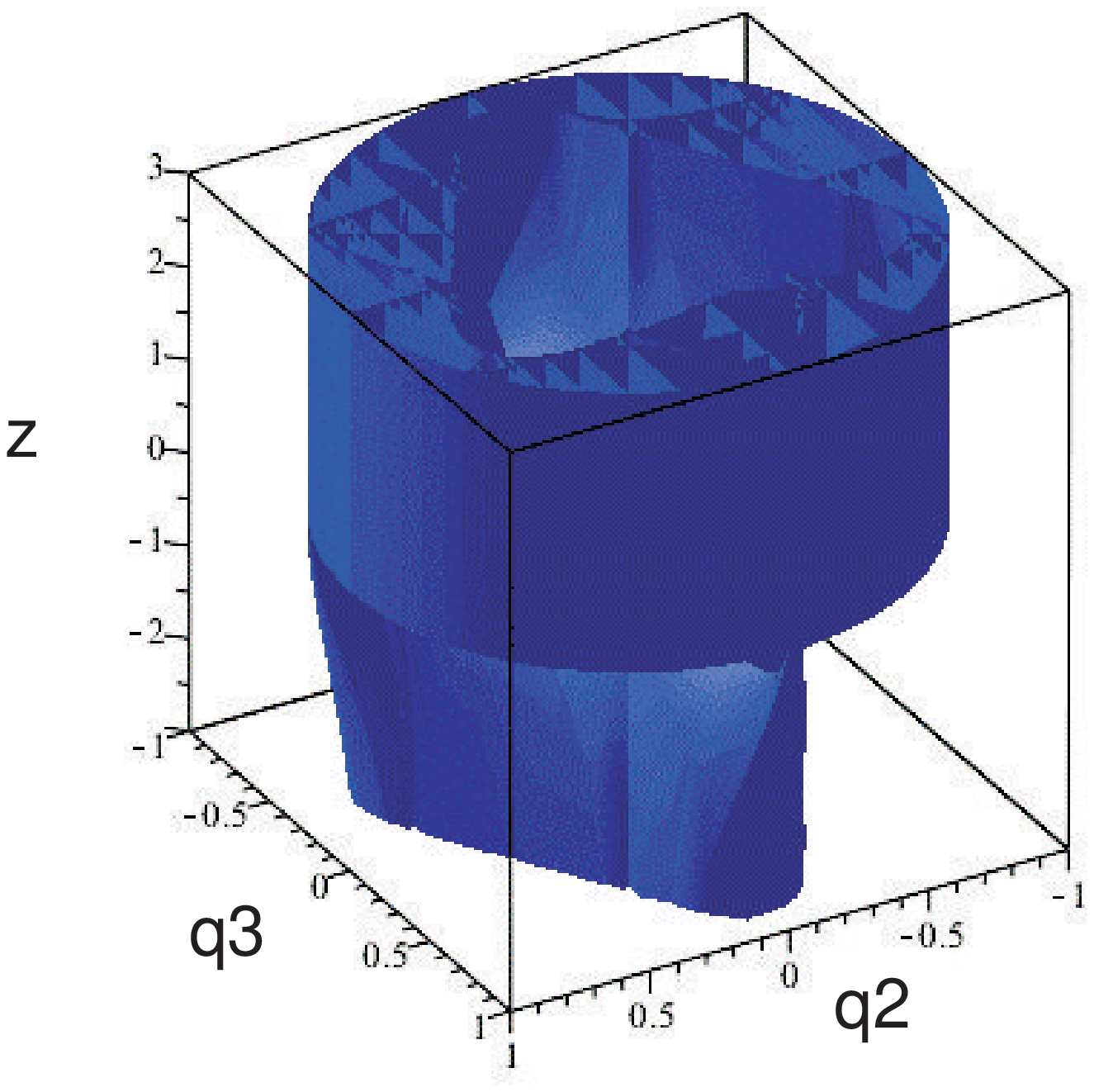}(d)
       \end{minipage}
    \end{tabular}
    \caption{Aspects for $OM_1$ with $\det({\bf A})<0$ (a) and $\det({\bf A})>0$ (b) and aspects for $OM_2$ with $\det({\bf A})<0$ (c) and $\det({\bf A})>0$ (d)}
    \protect\label{figure:Workspace_q4_q1}
    \end{center}
\end{figure}

In \cite{Wenger:1997}, the notion of aspect is defined for parallel robots with only one inverse kinematic solution. An aspect $WA_i$ is a maximal singularity free set defined such that:
\begin{itemize}
  \item $WA_i \subset W$
	\item $WA_i$ is connected
	\item $\forall X \in WA_i$, ${\rm~det({\bf A})}\neq 0$ and  ${\rm det({\bf B})}\neq 0$
\end{itemize}
This notion is now extended for a parallel robot with several  operation modes such that:
\begin{itemize}
  \item $WA_{ij} \subset W_{OM_j}$
	\item $WA_{ij}$ is connected
	\item $\forall X \in WA_{ij}$,${\rm~det({\bf A})}\neq 0$ and  ${\rm det({\bf B})}\neq 0$
\end{itemize}
In other words, an aspect $WA_{ij}$ is the maximum connected region without any singularity of the $OM_j$.  The analysis of the workspace is done in the projection space ($z$, $q_2$, $q_3$), and shows the existence of four aspects as shown in Fig.~\ref{figure:Workspace_q4_q1}. 
However, no further analysis is done to prove this feature in the four dimension space. 
As there are several solutions for the DKP in the same aspect, non-singular assembly mode trajectories are possible. The cylindrical algebraic decomposition (CAD) implemented in the SIROPA library has been used to decompose an aspect into a set of cells where algebraic equations define its boundaries \cite{Chablat:2011}. 
The CAD provides a formal decomposition of the parameter space in cells where the polynomials ${\rm~det({\bf A})} $ and ${\rm~det({\bf B})}$ have a constant sign\cite{Collins:1975} and the number of solutions for the DKP is constant.  
\subsection{Characteristic surfaces for an operation mode}
The notion of  {\it characteristic surface} was introduced in \cite{Wenger:1992} to define the uniqueness domains for serial Cuspidal robots. This definition was extended to parallel robots with one inverse kinematic solution in \cite{Wenger:1997} and with several inverse kinematic solutions in \cite{Chablat:2001}. In this paper, we  introduce this notion for a parallel robot with several operating modes.

\begin{figure}[hbt]
    \begin{center}
    \begin{tabular}{cc}
       \begin{minipage}[t]{57 mm}
                \psfrag{R2}{$\rho_2$}
                \psfrag{R3}{$\rho_3$}
								\psfrag{cusp}{Cusp points}
								\psfrag{8 solutions}{8 DKP solutions}
								\psfrag{4 solutions}{4 DKP solutions}
								\psfrag{Om1}{$OM_1$}
        \includegraphics[scale=0.22]{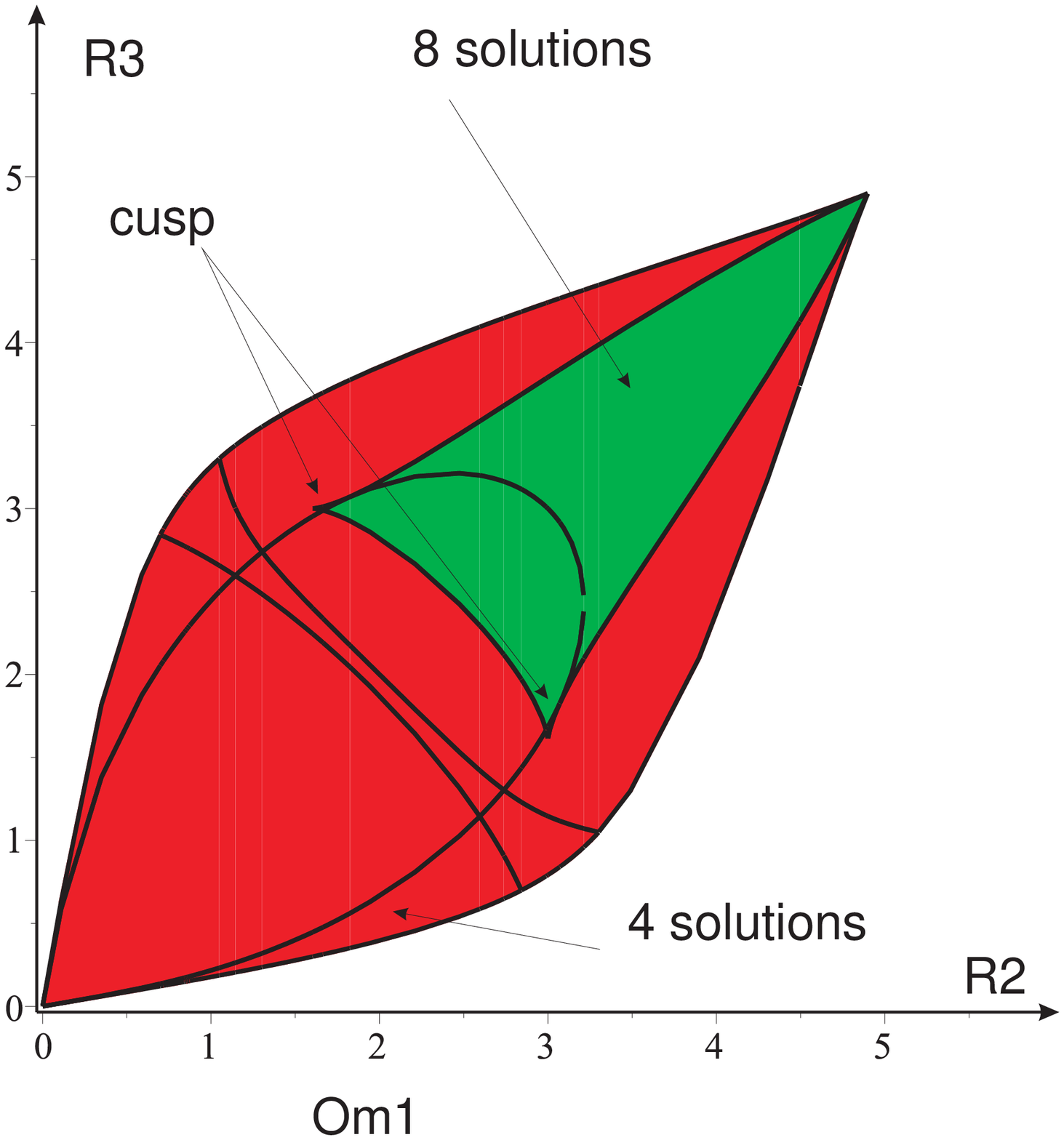}(a) 
       \end{minipage} &
		   \begin{minipage}[t]{57 mm}
                \psfrag{R2}{$\rho_2$}
                \psfrag{R3}{$\rho_3$}
								\psfrag{cusp}{Cusp points}
								\psfrag{8 solutions}{8 DKP solutions}
								\psfrag{4 solutions}{4 DKP solutions}
								\psfrag{Om2}{$OM_2$}
        \includegraphics[scale=0.22]{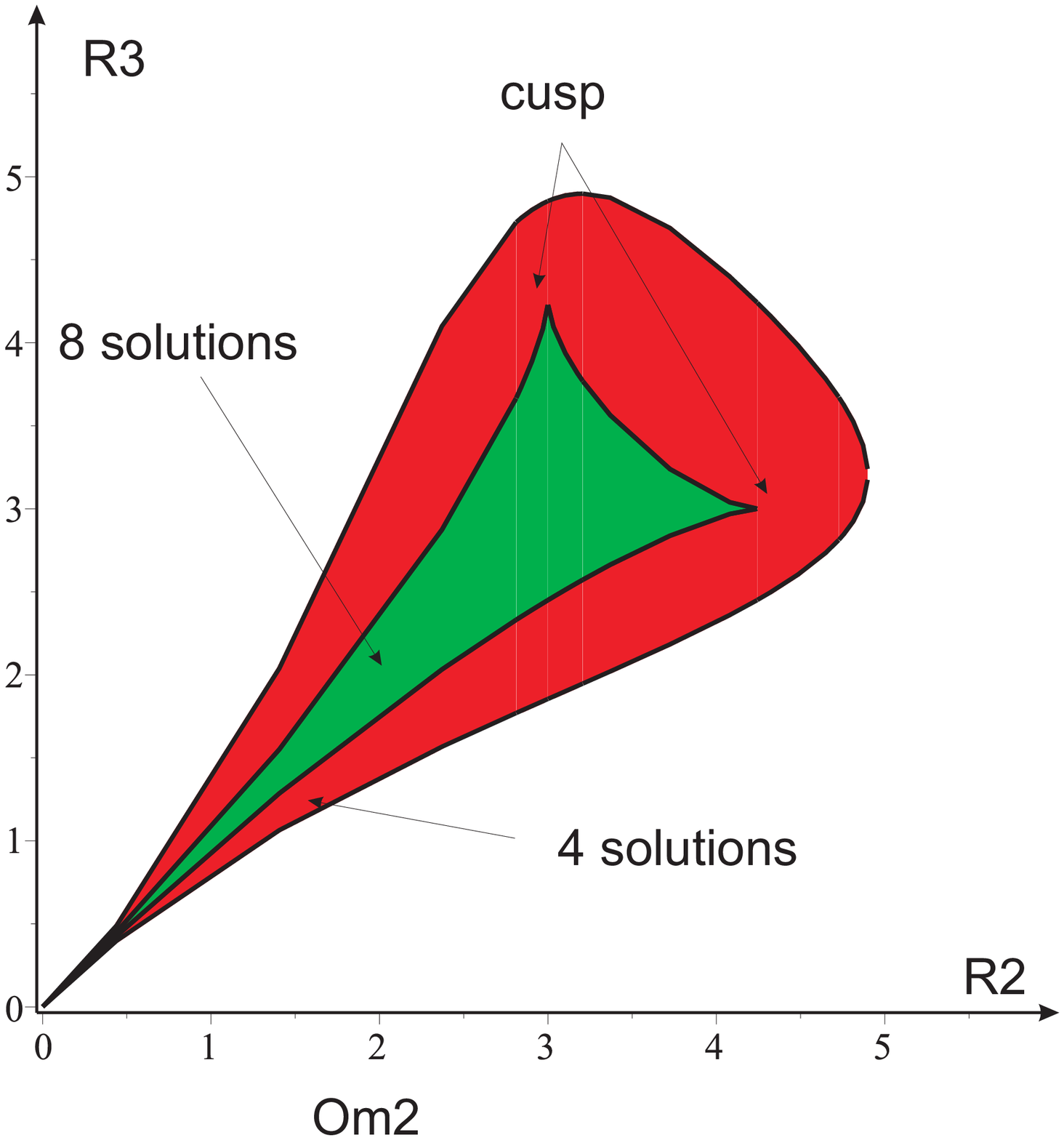}(b)
       \end{minipage} 
    \end{tabular}
    \end{center}
    \caption{Slice of the joint space for $\rho_1=3$ for $OM_1$ (a) and $OM_2$ (b)}
    \protect\label{figure:Jointspace_q4_q1}
\end{figure}

Let $WA_{ij}$ be one aspect for the operation mode $j$. The characteristic surfaces, denoted by ${\cal S}_C(WA_{ij})$, are defined as the preimage in  $WA_{ij}$ of the boundary $\overline {WA}_{ij}$  of  $WA_{ij}$.
\begin{equation}
{\cal S}_C(WA_{ij})= g_j^{-1}\left( g(\overline{WA_{ij}})\right) \cap WA_{ij}
\end{equation}
These characteristic surfaces are the images in the workspace of the singularity surfaces. By using the singularity and characteristic surfaces, we can compute the {\it basic regions} as defined in \cite{Wenger:1997}. The joint space is divided by the singularity surfaces in regions where the number of  solutions for the DKP is constant. We  also name these regions the {\it basic components} as in \cite{Wenger:1997}. For each operation mode, we find regions where the DKP admits four (in red) or eight (in green) solutions, as it is depicted in Fig.~\ref{figure:Jointspace_q4_q1}. We can also notice in Fig~\ref{figure:Jointspace_q4_q1} the existence of cusp points. 
\subsection{Non-singular assembly mode changing trajectories}
\begin{figure}[!hb]
    \begin{center}
    \begin{tabular}{cc}
       \begin{minipage}[t]{57 mm}
				\psfrag{q2}{$q_2$}
				\psfrag{q3}{$q_3$}
				\psfrag{P1}{$P_1$}
				\psfrag{P2}{$P_2$}
				\psfrag{P3}{$P_3$}
				\psfrag{S}{${\cal S}_{OM_1}$}
				\psfrag{SS}{${\cal S}_C$}
				\includegraphics[scale=0.22]{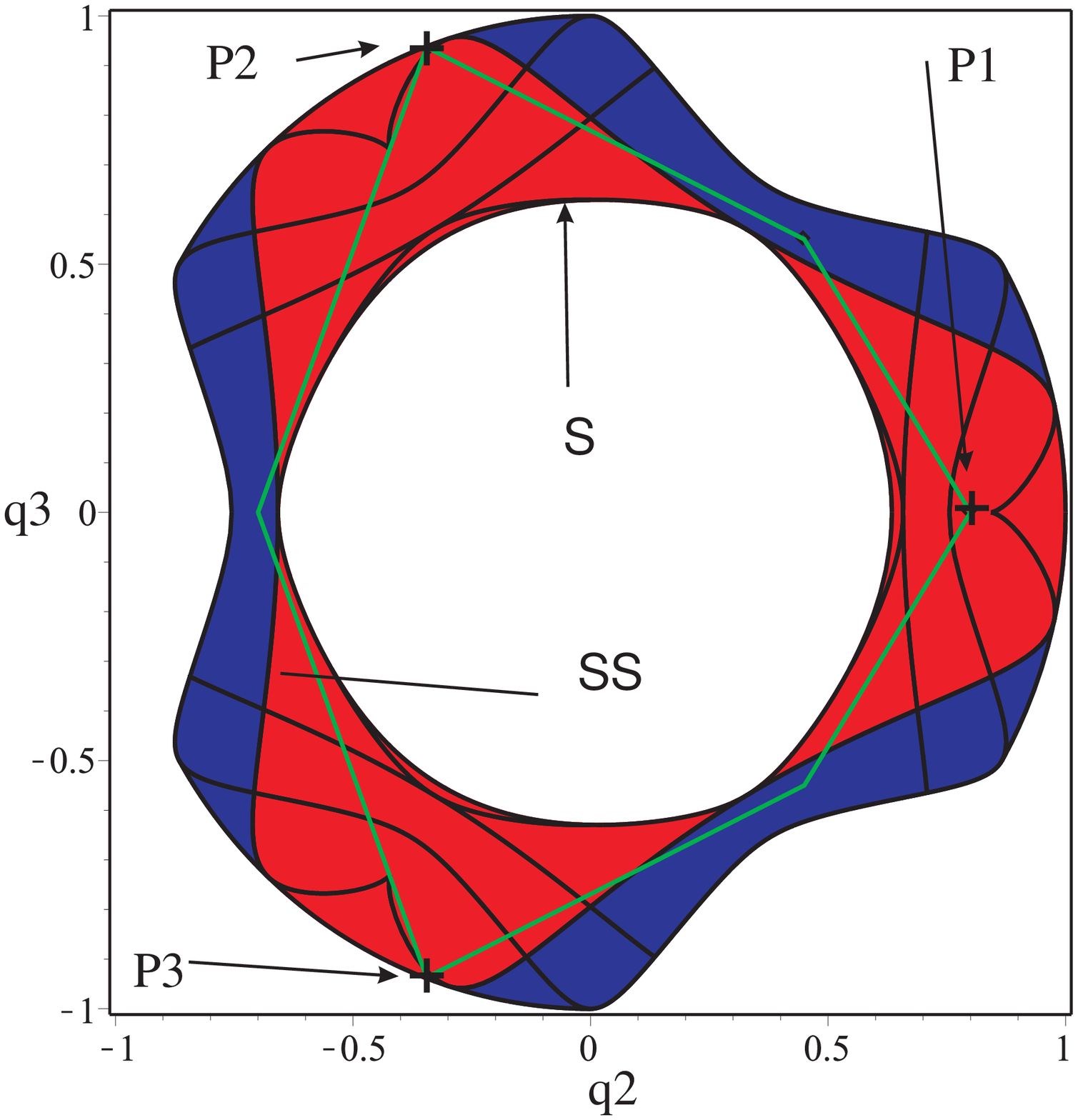} (a)
       \end{minipage} &
       \begin{minipage}[t]{57 mm}
				\psfrag{q2}{$q_2$}
				\psfrag{q3}{$q_3$}
				\psfrag{P1}{$P_5$}
				\psfrag{P2}{$P_6$}
				\psfrag{P3}{$P_7$}
				\psfrag{S}{${\cal S}_{OM_2}$}
				\psfrag{SS}{${\cal S}_C$}
				\includegraphics[scale=0.22]{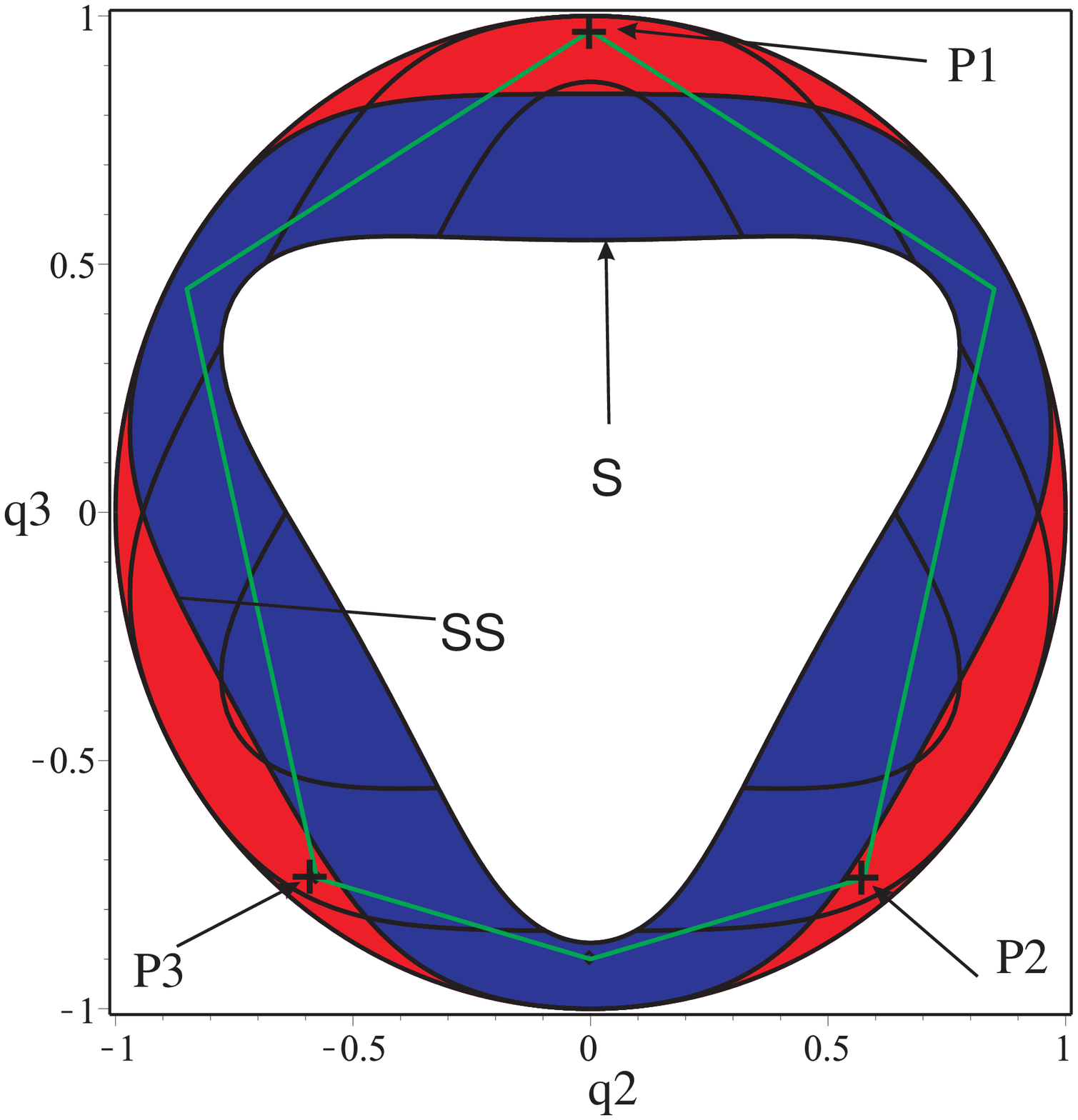}(b)
       \end{minipage}
    \end{tabular}
    \caption{Slice of an aspect for $z=3$ and $\det({\bf A})>0$ for $OM_1$ (a) and $OM_2$ (b) with in blue (resp. in red) a basic region coming from a basic component with four DKP (resp. eight)}
    \protect\label{figure:Workspace_q1_q4_p}
    \end{center}
\end{figure}

\begin{figure}[hbt]
    \begin{center}
    \begin{tabular}{cc}
       \begin{minipage}[t]{57 mm}
				\psfrag{R1}{$\rho_1$}
				\psfrag{R2}{$\rho_2$}
				\psfrag{R3}{$\rho_3$}
				\includegraphics[scale=0.22]{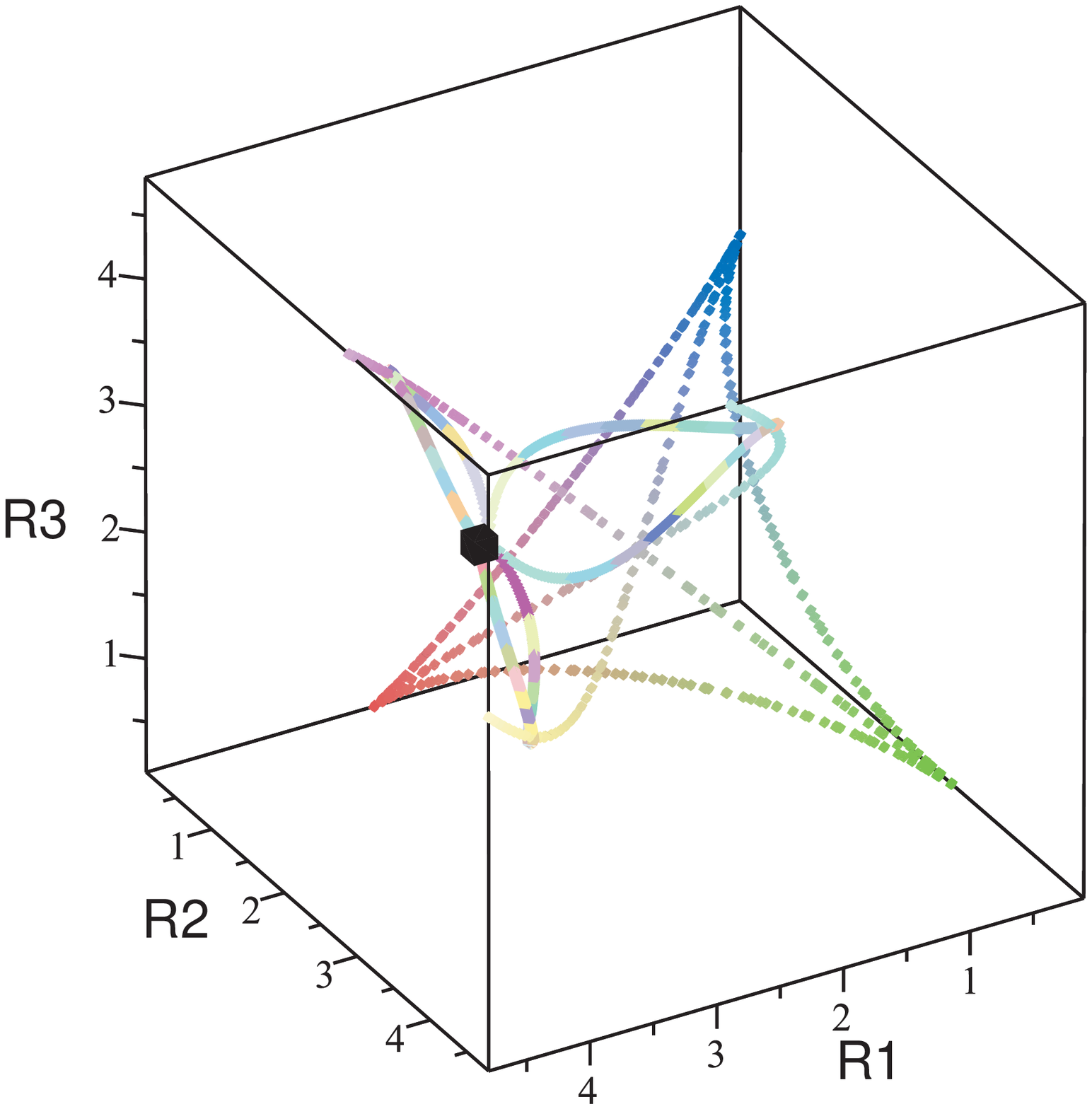}(a)
       \end{minipage} &
       \begin{minipage}[t]{57 mm}
				\psfrag{R1}{$\rho_1$}
				\psfrag{R2}{$\rho_2$}
				\psfrag{R3}{$\rho_3$}
				\includegraphics[scale=0.22]{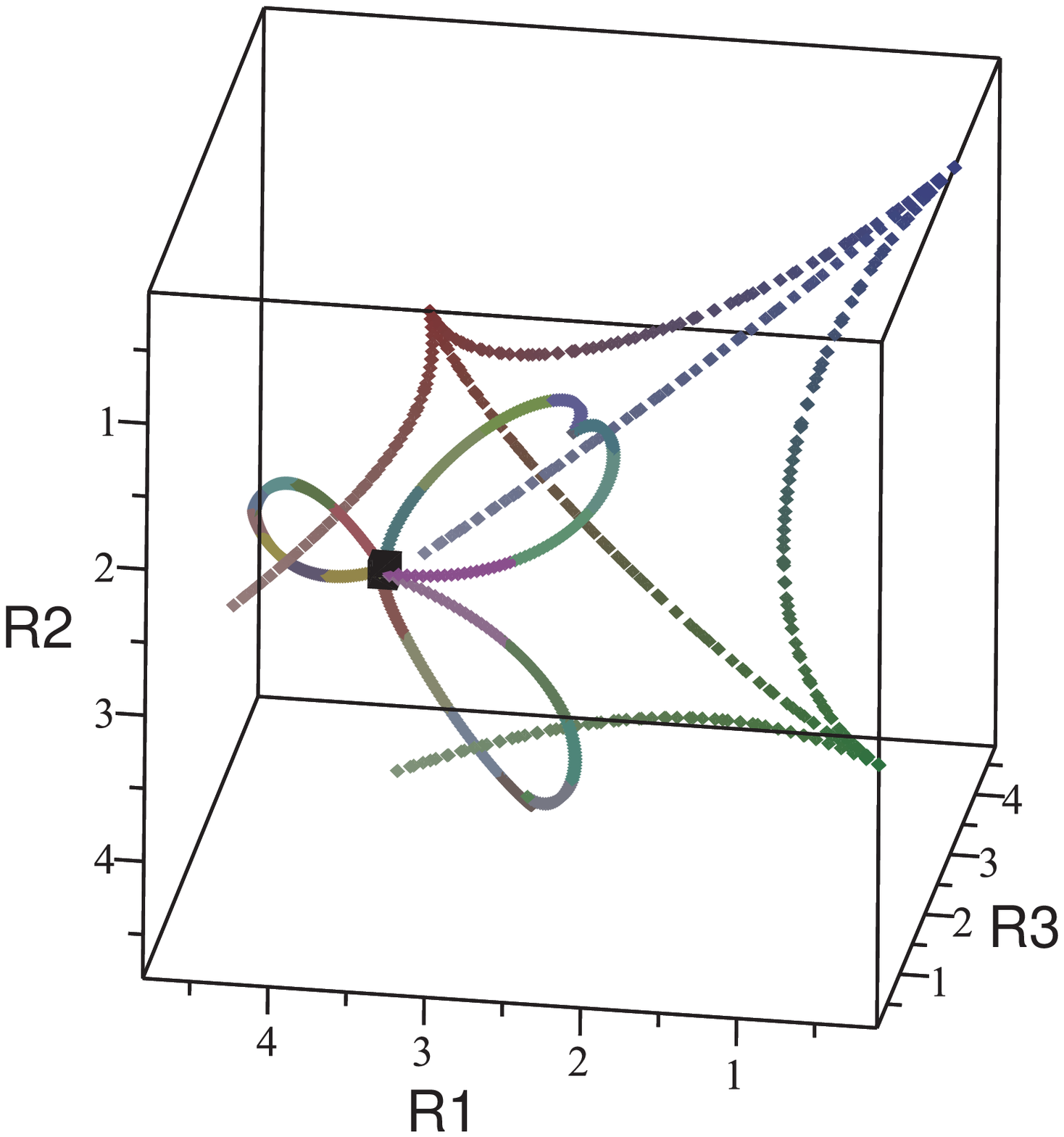} (b)
       \end{minipage}
    \end{tabular}
		\caption{Projection in $Q$ of the trajectories with the cusp curves for $OM_1$ (a) and $OM_2$ (b) }
    \protect\label{figure:jointspace_q1_q4}
    \end{center}
\end{figure}

Due to the lack of space and for pedagogical purpose, we only report a slice of the workspace. Letting $z=3$, the basic regions are computed by using the cylindrical algebraic decomposition for a given aspect. Figure~\ref{figure:Workspace_q1_q4_p} shows the three basic regions' image of basic components with 8 solutions for the DKP  
and a single basic region's image of a basic components with 4 solutions for the DKP connects these three previous basic regions. 
Table~\ref{table:DKP} presents the roots of the DKP for ${\rm det}(\bf A)>0$ for a joint position in each operation mode. For each of them, we find out that three roots have their $z$ coordinate close to $3$. A non-singular assembly mode changing trajectory can be obtained between three basic regions coming from eight solutions to the DKP. Due to symmetrical properties, there are also three roots of the DKP for ${\rm det}(\bf A)<0$   with $z=-3$. For $OM_1$, we construct a path between $P_1$, $P_2$, $P_3$ and for $OM_2$ between $P_5$, $P_6$ and $P_7$. When a straight line between two poses cross a singularity, we add an intermediate point as shown in Fig.~\ref{figure:Workspace_q1_q4_p}. The connections between the basic regions depicted in red are the projections of the cusp points in the workspace, i.e. the tangent between the singularity surface and the characteristic surface \cite{Chablat:2011}.
\begin{table} [h!]
\begin{center}
    \caption{Solutions of the DKP for ${det}(\bf A)>0$}
    \protect\label{table:DKP}
    \begin{tabular}{|c|c|c|c|c|c|c|c|c|c|}
    \hline
    \multicolumn{5}{|c|}{$OM_1$} & \multicolumn{5}{|c|}{$OM_2$}\\ \hline
		\multicolumn{5}{|c|}{$\rho_1 = 3.90$, $\rho_2 = 3.24$, $\rho_3 = 3.24$} & \multicolumn{5}{|c|}{$\rho_1 = 3.79$, $\rho_2 = 3.24$, $\rho_3 = 3.24$}\\ \hline
		$P$ & $z$ & $q_2$ & $q_3$ & $q_4$ & $P$ & $z$ & $q_1$ & $q_2$ & $q_3$ \\ \hline
		$P_1$ & $ 3.01$ & $-0.34$ & $-0.94$  & $ 0.06$  &   $P_5$ & $ 3.04$ & $ 0.35$ & $-0.58$  & $-0.74$ \\ \hline
		$P_2$ & $ 3.01$ & $-0.34$ & $ 0.94$  & $ 0.06$  &   $P_6$ & $ 3.04$ & $ 0.35$ & $ 0.586$ & $-0.74$ \\ \hline
		$P_3$ & $ 3$    & $ 0.85$ & $ 0.0$   & $ 0.53$  & 	$P_7$ & $ 3$    & $ 0.24$ & $ 0.0$   & $ 0.97$ \\ \hline
    $P_4$ & $-2.88$ & $-0.35$ & $ 0.0$   & $ 0.93$  &   $P_8$ & $-3.42$ & $ 0.98$ & $ 0.0$   & $ 0.19$ \\ \hline
    \hline
    \end{tabular}
\end{center}
\end{table}
\begin{figure}[hbt]
    \begin{center}
    \begin{tabular}{cc}
       \begin{minipage}[t]{57 mm}
				\psfrag{q2}{$q_2$}
				\psfrag{q3}{$q_3$}
				\psfrag{P1}{$P_1$}
				\psfrag{P2}{$P_2$}
				\psfrag{P3}{$P_3$}
				\psfrag{det}{${\rm det}({\bf A})$}
				\includegraphics[scale=0.192]{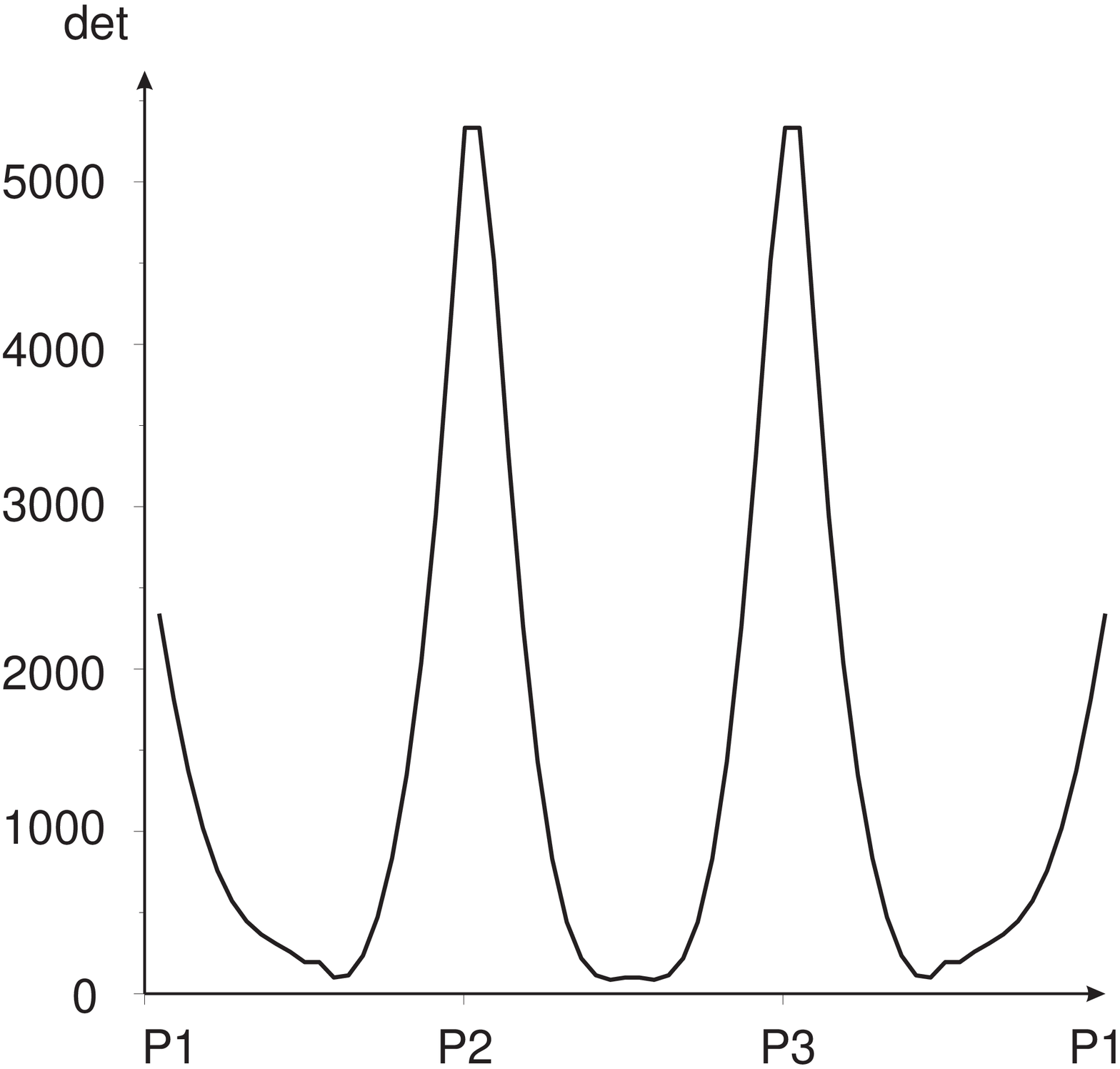}~~(a)
       \end{minipage} &
       \begin{minipage}[t]{57 mm}
				\psfrag{q2}{$q_2$}
				\psfrag{q3}{$q_3$}
				\psfrag{P1}{$P_5$}
				\psfrag{P2}{$P_6$}
				\psfrag{P3}{$P_7$}
				\psfrag{det}{${\rm det}({\bf A})$}
				\includegraphics[scale=0.192]{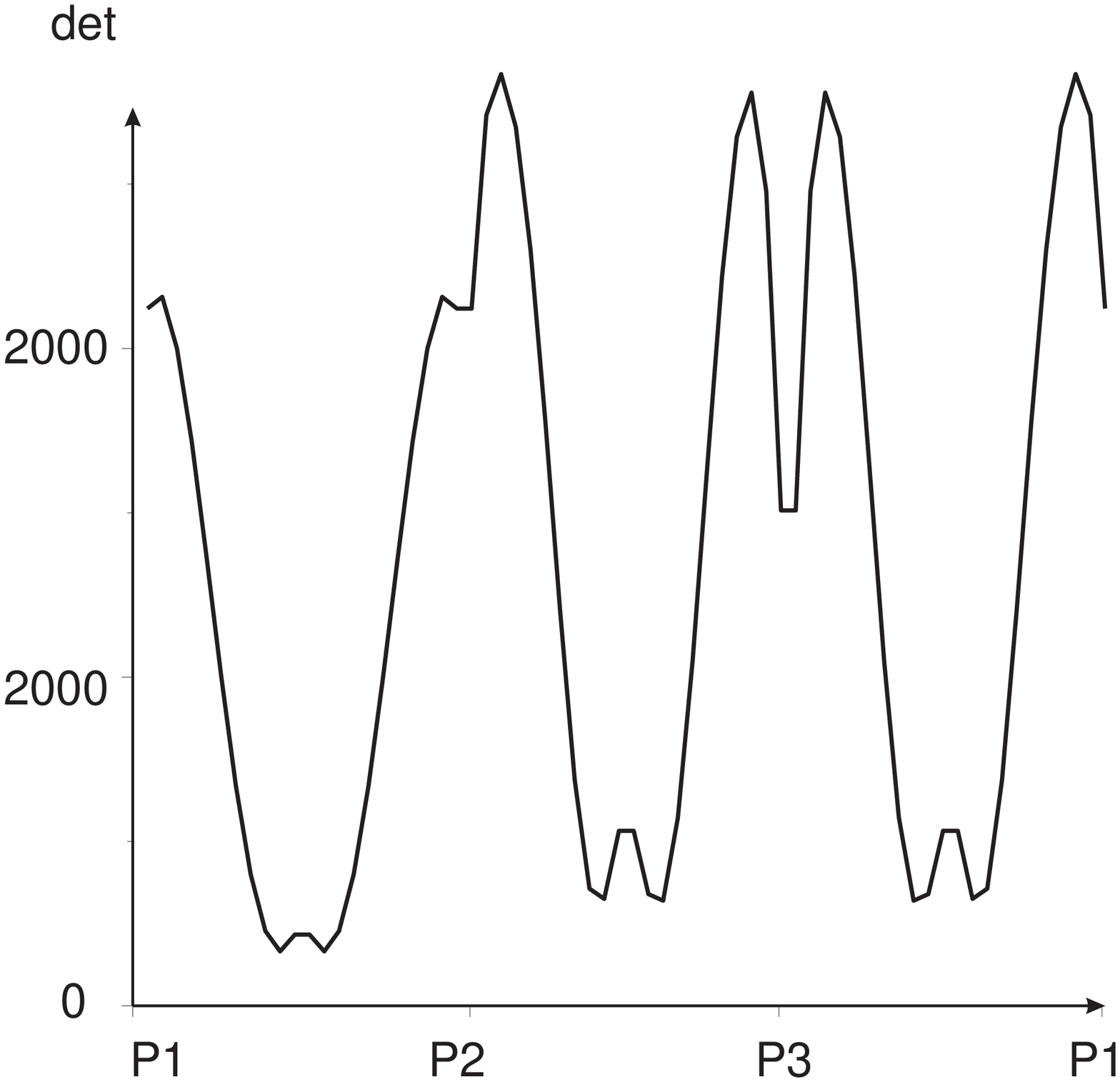}~~(b)
       \end{minipage}
    \end{tabular}
    \caption{Variation of $\det({\bf A})$ along trajectory $P_1, P_2, P_3$ for $OM_1$ (a) and $P_5, P_6, P_7$ for $OM_2$ (b)}
    \protect\label{figure:Workspace_q1_q4_det}
    \end{center}
\end{figure}
The variation of the ${\rm det}({\bf A})$ is plotted in the Figure~\ref{figure:Workspace_q1_q4_det}  and shows the existence of a non-singular assembly mode changing trajectory. The image of this trajectory in the joint space is illustrated in the Figure~\ref{figure:jointspace_q1_q4}. 
The projection of the cyclic trajectory defined by ($P_1$, $P_2$, $P_3$, $P_1$) (resp. ($P_5$, $P_6$, $P_7$, $P_5$)) onto the joint space encloses three curves of cusps. 
This behavior is similar to that of the  3-R\underline{P}R robot described in \cite{Zein:2008} or the 3-R\underline{P}S robot in \cite{Husty:2013}. The path to connect the fourth solutions is not presented in this paper. The method introduced in \cite{Moroz:2010} is used to compute the cusp curves.
\section{Conclusions}
\label{sec:conclusions}
This article presents a study of the joint space and workspace of the 3-R\underline{P}S parallel robot and shows the existence of non-singular assembly mode changing trajectories. 
First, we have shown that each of the two operation modes is divided into two aspects, which is a necessary condition for non-singular assembly mode changing trajectories. Moreover, it turns out that this mechanism has a maximum of 16 real solutions to the direct kinematic problem, eight for each operation mode.
Then, by computing the characteristic surfaces, we have shown that we can describe the basic regions for each operation mode. We construct a path going through several basic regions which are images of the same basic component with 8 solutions for the DKP.  The analysis of the determinant of Jacobian shows that a non-singular assembly mode change exists for each motion type. 
\bibliographystyle{unsrt}

\end{document}